\documentclass[letterpaper, 10 pt, conference]{ieeeconf}  

\IEEEoverridecommandlockouts                              

\usepackage{graphicx} 

\usepackage{cite}
\usepackage{algorithm}
\usepackage{algorithmic}
\usepackage{amsfonts}
\usepackage{amsmath}
\usepackage{array,booktabs,arydshln}
\usepackage{bbm}
\usepackage{subfig}

\usepackage{xcolor}

\definecolor{green}{RGB}{11,155,13}

\title{\LARGE \bf
APPLR: Adaptive Planner Parameter Learning from Reinforcement
}

\author{Zifan Xu$^{1}$, Gauraang Dhamankar$^{2}$, Anirudh Nair$^{3}$, Xuesu Xiao$^{2}$, \\Garrett Warnell$^{4}$, Bo Liu$^{2}$, Zizhao Wang$^{5}$, and Peter Stone$^{2,6}$ 
\thanks{Department of $^{1}$Physics {\tt\small zfxu@utexas.edu}, 
        $^{2}$Computer Science {\tt\small \{dgauraang, xiao, bliu, pstone\}@cs.utexa.edu}, 
        $^{3}$Mathematics {\tt\small ani.nair@utexas.edu}, 
        $^{5}$Electrical and Computer Engineering {\tt\small zizhao.wang@utexas.edu},
        University of Texas at Austin, Austin, Texas 78712.
        $^{4}$Computational and Information Sciences Directorate, Army Research Laboratory, Adelphi, MD 20783 {\tt\small garrett.a.warnell.civ@mail.mil}.
        $^{6}$Sony AI.
        This work has taken place in the Learning Agents Research
Group (LARG) at UT Austin.  LARG research is supported in part by NSF
(CPS-1739964, IIS-1724157, NRI-1925082), ONR (N00014-18-2243), FLI
(RFP2-000), ARO (W911NF-19-2-0333), DARPA, Lockheed Martin, GM, and
Bosch.  Peter Stone serves as the Executive Director of Sony AI
America and receives financial compensation for this work.  The terms
of this arrangement have been reviewed and approved by the University
of Texas at Austin in accordance with its policy on objectivity in
research.}%
}

\begin{document}

\maketitle
\thispagestyle{empty}
\pagestyle{empty}

\begin{abstract}
Classical navigation systems typically operate using a fixed set of hand-picked parameters (e.g. maximum speed, sampling rate, inflation radius, etc.) and require heavy expert re-tuning in order to work in new environments. To mitigate this requirement, it has been proposed to learn parameters for different contexts in a new environment using human demonstrations collected via teleoperation. 
However, learning from human demonstration limits deployment to the training environment, and limits overall performance to that of a potentially-suboptimal demonstrator. 
In this paper, we introduce \textsc{applr}, \emph{Adaptive Planner Parameter Learning from Reinforcement}, which allows existing navigation systems to adapt to new scenarios by using a parameter selection scheme discovered via reinforcement learning (RL) in a wide variety of simulation environments. We evaluate \textsc{applr} on a robot in both simulated and physical experiments, and show that it can outperform both a fixed set of hand-tuned parameters and also a dynamic parameter tuning scheme learned from human demonstration.
\end{abstract}

\section{INTRODUCTION}
\label{sec::intro}

Most classical autonomous navigation systems are capable of moving robots from one point to another, often with verifiable collision-free guarantees, under a set of parameters (e.g. maximum speed, sampling rate, inflation radius, etc.) that have been fine-tuned for the deployment environment.
However, these parameters need to be re-tuned to adapt to different environments, which requires extra time and energy spent onsite during deployment, and more importantly, expert knowledge of the inner workings of the underlying navigation system~\cite{zheng2017ros, xiao2017uav}.

A recent thrust to alleviate the costs associated with expert re-tuning in new environments is to learn an adaptive parameter tuning scheme from demonstration~\cite{xiao2020appld}. Although this approach removes the requirement of expert tuning, it still depends on access to a human demonstration, and the learned parameters are typically only applicable to the training environment. Moreover, the performance of the system is limited by the quality of the human demonstration, which may be suboptimal.

\begin{figure}
  \centering
  \includegraphics[width=\columnwidth]{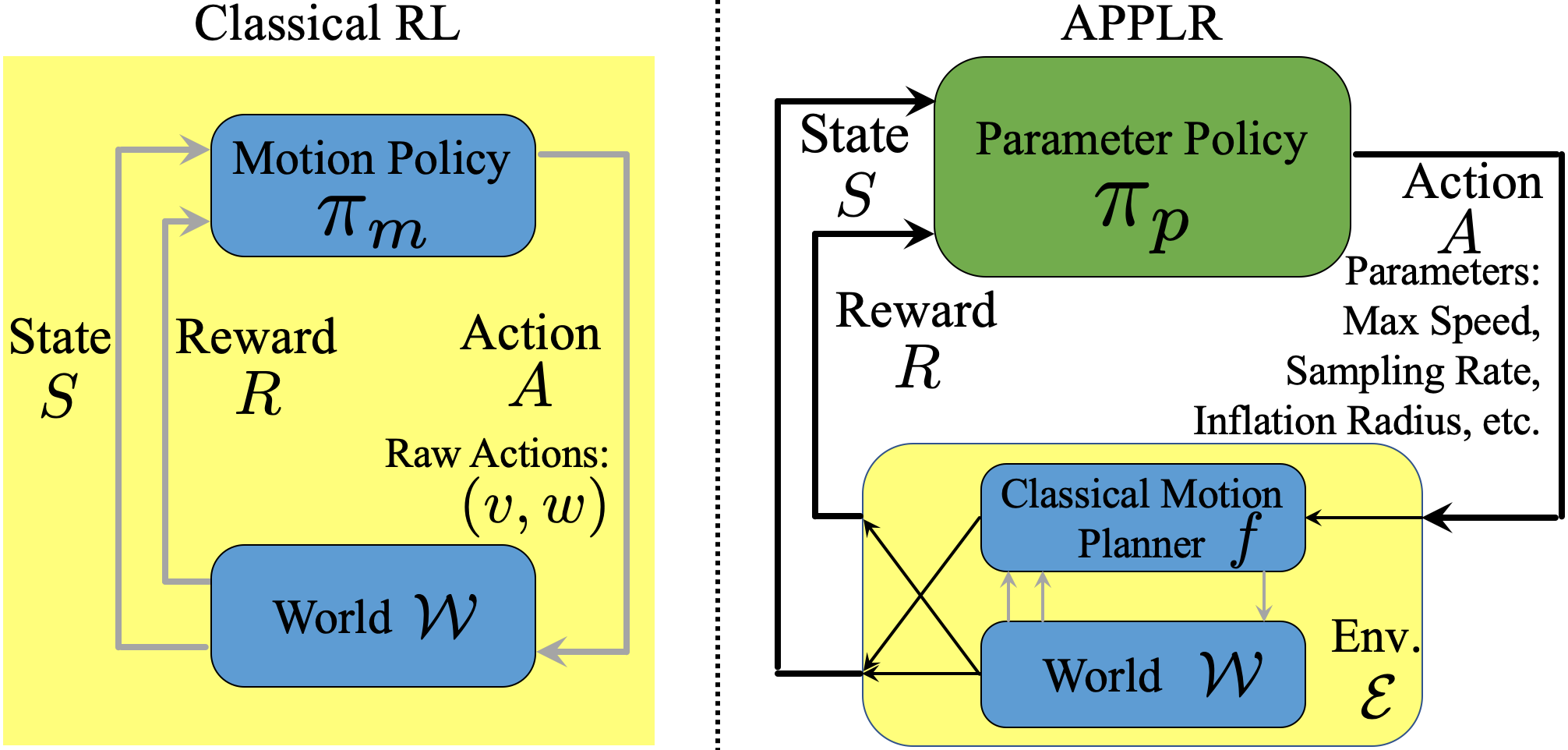}
  \caption{Instead of learning an end-to-end motion policy $\pi_m$ which takes state $S$ and reward $R$ from the world $\mathcal{W}$ and produces raw actions $A$, e.g. linear and angular velocity $(v, \omega)$ (left), \textsc{applr} treats an underlying classical motion planner $f$ as part of the \emph{meta-environment} $\mathcal{E}$ (along with the world $\mathcal{W}$) and the learned \emph{parameter policy} $\pi_p$ interacts with it through actions in the \emph{parameter} space (right). 
  In this way, the RL agent selects its action in the form of a set of navigation parameters at \emph{each time step} and reasons about potential \emph{future} consequences of those parameters, rather than \emph{tuning} a single set of parameters for the \emph{entire environment} only considering the \emph{current} situation. 
  }
  \label{fig::applr}
\end{figure}

In this paper, we seek a new method for adaptive autonomous navigation which does \emph{not} need access to expert tuning or human demonstration, and is generalizable to many deployment environments. 
We hypothesize that a method based on reinforcement learning in simulation could achieve these goals, and we verify this hypothesis by proposing and studying \emph{Adaptive Planner Parameter Learning from Reinforcement} (\textsc{applr}). 
By using reinforcement learning, \textsc{applr} introduces the concept of a \emph{parameter policy} (Fig. \ref{fig::applr}), which is trained to make planner parameter decisions in such a way that allows the system to take suboptimal actions at one state in order to perhaps perform even better in the future. For example, while it may be suboptimal in the moment to slow down or alter the platform's trajectory before a turn, doing so may allow the system to carefully position itself so that it can go much faster in the future than if it had not.
Additionally, as opposed to an end-to-end motion policy (i.e., a mapping from states to low-level motion commands), \textsc{applr}'s parameter policy interacts with an underlying classical motion planner, and therefore the overall system inherits all the benefits enjoyed by classical approaches (e.g., safety and explainability). We posit that learning policies that act in the parameter space of an existing motion planner instead of in the velocity control space can increase exploration safety, improve learning efficiency, generalize well to unseen environments, and allow effective sim-to-real transfer.

\section{RELATED WORK}
\label{sec::related}

In this section, we summarize related work on existing parameter tuning approaches for classical navigation and on learning-based navigation systems. 

\subsection{Parameter Tuning}
Classical navigation systems usually operate under a static set of parameters. Those parameters are manually adjusted to near-optimal values based on the specific deployment environment, such as high sampling rate for cluttered environments or high maximum speed for open space. This process is commonly known as \emph{parameter tuning}, which requires robotics experts' intuition, experience, or trial-and-error~\cite{zheng2017ros, xiao2017uav}. 
To alleviate the burden of expert tuning, automatic tuning systems have been proposed, such as those using fuzzy logic~\cite{teso2019predictive} or gradient descent~\cite{bhardwaj2019differentiable}, to find \emph{one} set of parameters tailored to the specific navigation scenario. Recently, Xiao {\em et al.} introduced Adaptive Planner Parameter Learning from Demonstration (\textsc{appld}), which learns a \emph{library} of parameter sets for different navigation \emph{contexts} from teleoperated demonstration, and dynamically tunes the underlying navigation system during deployment. \textsc{appld}'s parameter-tuning ``on-the-fly" opens up a new possibility for improving classical navigation systems.
However, \textsc{appld} makes decisions about which parameters to use based exclusively on the current context in a manner that disregards the potential future consequences of those decisions.

In contrast, \textsc{applr} goes beyond this myopic parameter tuning scheme and introduces the concept of a parameter \emph{policy}, where we use the term policy to explicitly denote state-action mappings found through reinforcement learning to solve long-horizon sequential decision making problems. By using such a policy, \textsc{applr} is able to make parameter-selection decisions that take into account the possible future consequences of those decisions.

\subsection{Learning-based Navigation}
A plethora of recent works have applied machine learning techniques to the classical mobile robot navigation problem~\cite{pfeiffer2017perception, gao2017intention, chiang2019learning, xiao2020toward, liu2020lifelong, xiao2020agile}. By directly leveraging experiential data, these learning approaches can not only enable point-to-point collision-free navigation without sophisticated engineering, but also enable capabilities such as terrain-aware navigation~\cite{siva2019robot, kahn2020badgr} and social navigation~\cite{everett2018motion, pokle2019deep}. However, most end-to-end learning approaches are data-hungry, requiring hours of training time and millions of training data/steps, either from expert demonstration (as in imitation learning) or trial-and-error exploration (RL). Moreover, when an end-to-end policy is trained in simulation using RL, the sim-to-real gap may cause problems in the real-world. Most importantly, learning-based methods typically lack safety and explainability, both of which are important properties for mobile robots interacting with the real-world.

\textsc{applr} aims to address the aforementioned shortcomings: as a RL approach, learning in parameter space (instead of in velocity control space) effectively eliminates catastrophic failures (e.g. collisions) and largely reduces costly random exploration, with the help of the underlying navigation system. This change in action space can also help to generalize well to unseen environments and to mitigate the difference between simulation and the real-world (e.g. physics).

\section{APPROACH}
\label{sec::approach}
We now introduced the proposed approach, \textsc{applr}, which aims to apply RL to identify an optimal parameter selection policy for a classical motion planner. By doing so, \textsc{applr} naturally inherits from the classical planner its safety guarantees and ability to generalize to unseen environments. In addition, through RL, \textsc{applr} learns to autonomously and adaptively switch planner parameters ``on-the-fly'' and in a manner that considers future consequences without any expert tuning or human demonstration. In the rest of this section, we first introduce the background of classical motion planning. Then, we provide the problem definition of \textsc{applr} under the standard Markov Decision Process framework. Finally, we discuss the designed reward functions and the chosen reinforcement learning algorithm.

\subsection{Background on Motion Planning}
In this work, we assume the robot employs a classical motion planner, $f$, that can be tuned through a set of planner parameters $\theta \in \Theta$. While navigating in a physical world $\mathcal{W}$,\footnote{In classical RL approaches for navigation, $\mathcal{W}$ is usually defined as an MDP itself with the conventional state and action space (Fig. \ref{fig::applr}).} $f$ tries to move the platform to a global navigation goal, e.g., $\beta = (\beta_x, \beta_y) \in \mathbb{R}^2$. At each time step $t$, $f$ receives sensor observations $o_t$ (e.g. lidar scans), and then computes a local goal $g=(g_x, g_y) \in \mathbb{R}^2$, which the robot attempts to reach quickly. Then, $f$ is responsible for computing the motion commands $u_t = f(o_t, \beta~|~\theta)$ (e.g. $u_t$ can be the angular/linear velocity). Most prior work in the learning community attempts to replace $f$ entirely by a learnable function $\pi_m$ that directly outputs the motion commands. However, the performance of these end-to-end planners is usually limited due to insufficient training data and poor generalization to unseen environments. In contrast, we focus here on a scheme for adjusting the planner parameters $\theta$ ``on-the-fly''. We expect learning in the planner parameter space to increase the overall system's adaptability while still benefiting from the verifiable safety assurance provided by the classical system.

\subsection{Problem Definition}
We formulate the navigation problem as a Markov Decision Process (MDP), i.e., a tuple $(\mathcal{S}, \mathcal{A}, \mathcal{T}, \gamma, R)$. 
Assume an agent is located at the state $s_t \in \mathcal{S}$ at time step $t$. If the agent executes an action $a_t \in \mathcal{A}$, the environment will transition the agent to $s_{t+1} \sim \mathcal{T}(\cdot | s_t, a_t)$ and the agent will receive a reward $r_t = R(s_t, a_t)$. The overall objective of RL is to learn a policy $\pi: \mathcal{S} \rightarrow \mathcal{A}$ that can be used to select actions that maximize the expected cumulative reward over time, i.e. $J = \mathbb{E}_{(s_t, a_t) \sim \pi}[\sum_{t=0}^\infty \gamma^t r_t]$.

In \textsc{applr}, we seek a policy in the context of an MDP $\mathcal{E}$ that denotes a \emph{meta-environment} composed of both the underlying navigation world $\mathcal{W}$ (the physical, obstacle-occupied world) and a given motion planner $f$ with adjustable parameters $\theta \in \Theta$ and sensory inputs $o \in \mathcal{O}$. 
Additionally, we assume going forward that a local goal $g$ is always available (as a waypoint along a coarse global path in most classical navigation systems), and we use its angle relative to the orientation of the agent, i.e., $\phi = \arctan2{(g_y, g_x)} \in [-\pi, \pi]$, to inject the local goal information to the agent.
%
Within $\mathcal{E}$, at each time step $t$, $s_t = (o_t, \phi_t, \theta_{t-1})$, where $o_t \in \mathcal{O}$ is the current sensory inputs, $\phi_t \in \mathcal{G}$ is the angle towards the local goal, and $\theta_{t-1} \in \Theta$ is the previous planner parameter that $f$ was using. 
That is, for our MDP $\mathcal{E}$, $\mathcal{S} = \mathcal{O} \times \mathcal{G} \times \Theta$ and $\mathcal{A} = \Theta$. 
In this context, \textsc{applr} aims to learn a policy $\pi_p:\mathcal{O} \times \mathcal{G} \times \Theta \rightarrow \Theta$ 
that selects a planner parameter $\theta_t$ that enables $f$ to achieve optimal navigation performance over time. The agent then transitions to the next state $s_{t+1} = (o_{t+1}, \phi_{t+1}, \theta_t)$, where $o_{t+1}, \phi_{t+1} \sim \mathcal{T}(\cdot | s_t, \theta_t)$ are given by $\mathcal{E}$. The overall objective is therefore
\begin{equation}
    \max_\pi J^\pi = \mathbb{E}_{s_0, \theta_t \sim \pi(s_t), s_{t+1} \sim \mathcal{T}(s_t,\theta_t)}\bigg[\sum_{t=0}^\infty \gamma^t r_t\bigg].
\end{equation}

After training with a reward function (Sec. \ref{sec::reward} and \ref{sec::rl}), the learned parameter policy $\pi_p$ is deployed with the underlying navigation system $f$ in the world $\mathcal{W}$, as summarized in Alg. \ref{alg::navigation}.


\begin{algorithm} 
		\caption{Navigation with \textsc{applr}}
		\label{alg::navigation}
		\begin{algorithmic}[1]
		    \REQUIRE{the physical world $\mathcal{W}$, the underlying motion planner $f$, the global goal $\beta$, the initial parameter $\theta_0$, and the parameter policy $\pi_p$.}
		    \STATE{$t = 1$} 
		    \WHILE{$\beta$ is not reached}
		    \STATE receive sensor readings $o_t$ from $\mathcal{W}$ and local goal $\phi_t$ from a coarse global plan
		    \STATE{$\theta_{t} = \pi_p(o_t, \phi_t, \theta_{t-1})$} \hfill \COMMENT{parameter policy}
		    \STATE $u_t = f(o_t, \beta~|~\theta_t)$ \hfill \COMMENT{$f$ parameterized by $\theta_t$} 
		    \STATE execute $u_t$ in $\mathcal{W}$
		    \STATE $t = t + 1$
		    \ENDWHILE
		\end{algorithmic}
\end{algorithm}

\subsection{Reward Function} 
\label{sec::reward}
We now describe our design of the reward function for \textsc{applr}. In general, we encourage three types of behaviors: (1) behaviors that lead to the global goal faster; (2) behaviors that make more local progress; and (3) behaviors that avoid collisions and danger. Correspondingly, the designed reward function can be summarized as
\begin{equation}
    R_t(s_t, a_t, s_{t+1}) = c_fR_f + c_p R_p + c_c R_c.
\end{equation}
Here, $c_f, c_p, c_c$ are coefficients for the three types of reward functions $R_f, R_p, R_c: \mathcal{S}\times \mathcal{A} \rightarrow \mathbb{R}$.
Specifically, $R_f(s_t, a_t) = \mathbbm{1}(s_t~\text{is terminal}) - 1$ applies a $-1$ penalty to every step before reaching the global goal.
To encourage the local progress of the robot, we add a dense shaping reward $R_p$. Assume at time $t$, the absolute coordinates of the robot are $p_t= (p^x_t, p^y_t)$, then we define
\begin{equation}
    R_p = \frac{(p_{t+1}-p_t)\cdot(\beta-p_t)}{|\beta-p_t|}.
\end{equation}
In other words, $R_p$ denotes the robot's local progress $(p_{t+1}-p_t)$ projected on the direction toward the global goal ($\beta - p_t$). 
Finally, a penalty for the robot colliding with or coming too close to obstacles is defined as $R_c = -1/d(o_{t+1})$, where $d(o_{t+1})$ is a distance function measuring how close the robot is to obstacles based on sensor observations. 

\subsection{Reinforcement Learning Algorithm} 
\label{sec::rl}
We consider two major factors for choosing the RL algorithm for \textsc{applr}: (1) the algorithm should allow selection of continuous actions since the parameter space of most planners is continuous; (2) the algorithm should be highly sample efficient for physical simulation of navigation. Based on these two criteria, we use the Twin Delayed Deep Deterministic policy gradient algorithm (TD3) \cite{fujimoto2018addressing} for \textsc{applr}. As one of the state-of-the-art off-policy algorithms, TD3 is very sample efficient and handles continuous actions by design.
Specifically, TD3 is an actor-critic algorithm that keeps an estimate for both the policy $\pi_p^\xi$ and the state-action value function $Q_p^\zeta$, parameterized by $\xi$ and $\zeta$ separately. For the policy, it uses the usual deterministic policy gradient update~\cite{silver2014deterministic}
\begin{equation}
    \label{eq:pg}
    \nabla_\xi J^{\pi_p^\xi} = \mathbb{E}_{s\sim \pi_p^\xi}\bigg[\nabla_a Q(s, a)|_{a=\pi_p^\xi} \nabla \pi^\xi_p(s)\bigg].
\end{equation}
To address the maximization bias on the estimation of $Q_p^\zeta$, which can influence the gradient in equation~\eqref{eq:pg}, TD3 borrows the idea from double $Q$ learning~\cite{hasselt2010double} of keeping two separate $Q$ estimators $Q_p^{\zeta_1}$ and $Q_p^{\zeta_2}$, each updated using the conventional Bellman residual objective $\mathbb{E}_{(s, a, r, s') \sim \mathcal{E}} \left[ ||Q_p^\zeta(s, a) - r - \gamma \max_{a'}Q_p^\zeta(s', a')||_2 \right]$. TD3 further stabilizes training by delaying the update of the critic until the value estimation is small and by using the clipped value $\min(Q_p^{\zeta_1}(s, a), Q_p^{\zeta_2}(s,a))$ in the place of the critic in equation~\eqref{eq:pg} to reduce overestimation of the true value. As physical simulations suffer from high variance and are generally slow, TD3 is a good fit for \textsc{applr}.

To further address the sample inefficiency issue, a distributed general reinforcement learning architecture (Gorila) \cite{nair2015massively} is employed, which enables parallelized acting processes on a computing cluster. Our implementation of Gorila is a simpler version with only one serial learner and a large number of actors running individually in simulation environments to generate large quantities of data for a global replay buffer.
\section{EXPERIMENTS}
\label{sec::experiments}

In this section, we experimentally validate that \textsc{applr} can enable adaptive autonomous navigation \emph{without} access to expert tuning or human demonstration and is generalizable to many deployment environments, both in simulation and in the real-world. 
To perform this validation, \textsc{applr} is implemented and applied on a Clearpath Jackal ground robot. 
The results of \textsc{applr} are compared with those obtained by the underlying navigation system using its default parameters from the robot platform manufacturer. For the physical experiments, we also compare to parameters learned from human demonstration using \textsc{appld}~\cite{xiao2020appld}. 

\subsection{Implementation}
We implement \textsc{applr} on a ClearPath Jackal differential-drive ground robot. The robot is equipped with a 720-dimensional planar laser scan with a 270$^\circ$ field of view, which is used as our sensory input $o_t$ in Alg. \ref{alg::navigation}. We preprocess the LiDAR scans by capping the maximum range to 2m.
The onboard Robot Operating System (ROS) \texttt{move\textunderscore base} navigation stack (our underlying classical motion planner $f$) uses Dijkstra's algorithm to plan a global path and runs \textsc{dwa}~\cite{fox1997dynamic} as the local planner. Our $u_t$ is the linear and angular velocity $(v, \omega)$. We query a local goal from the global path 1m  away  from  the  robot and compute the relative angle $\phi_t$. \textsc{applr} learns a parameter policy to select \textsc{dwa} parameters $\theta$, including \emph{max\_vel\_x}, \emph{max\_vel\_theta}, \emph{vx\_samples}, \emph{vtheta\_samples}, \emph{occdist\_scale}, \emph{pdist\_scale}, \emph{gdist\_scale}, and \emph{inflation\_radius}. We use the ROS \texttt{dynamic\textunderscore reconfigure} client to dynamically change planner parameters. The global goal $\beta$ is assigned manually, while $\theta_0$ is the default set of parameters provided by the robot manufacturer. 

\begin{figure}
  \centering
  \includegraphics[width=\columnwidth]{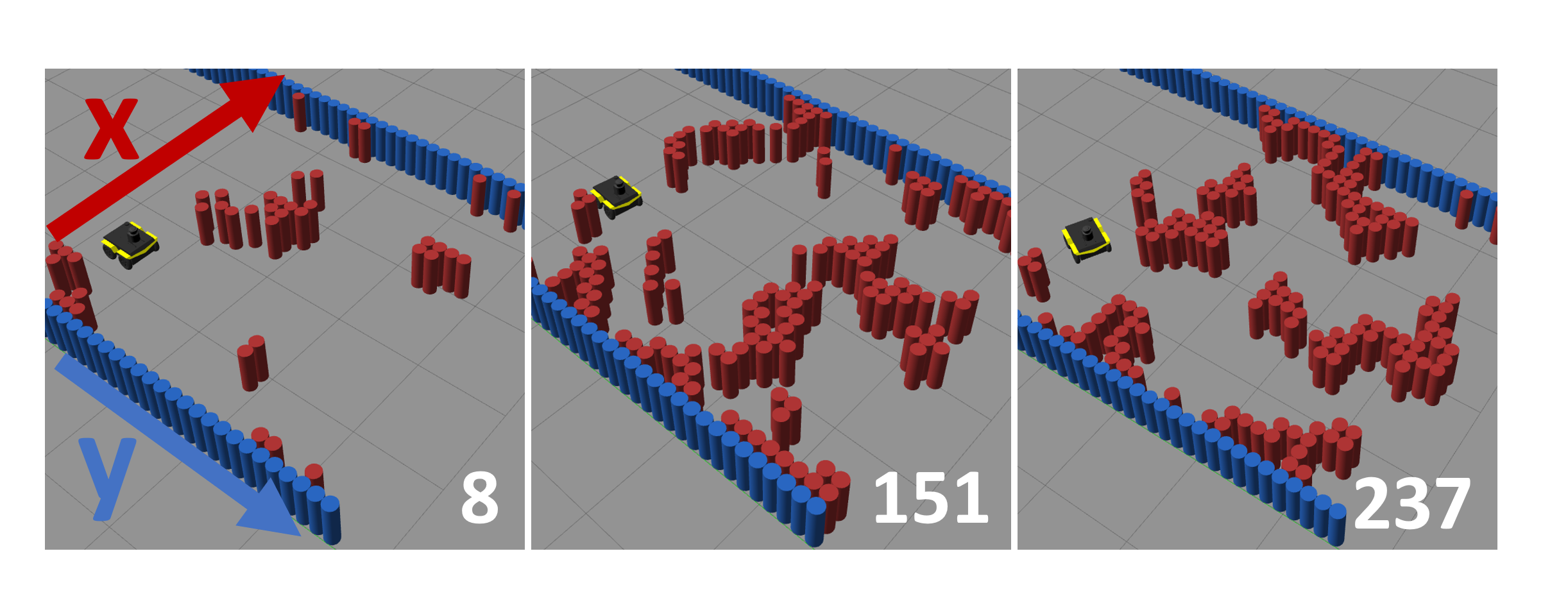}
  \caption{Indexed Example Navigation Environments in Benchmark for Autonomous Robot Navigation (BARN)~\cite{perille2020benchmarking}}
  \label{fig::barn}
\end{figure}

$\pi_p$ is trained in simulation using the Benchmark for Autonomous Robot Navigation (BARN) dataset~\cite{perille2020benchmarking} with 300 simulated navigation environments generated by cellular automata. Those environments cover different navigation difficulty levels, ranging from relatively open environments to highly-constrained spaces where the robot needs to squeeze through tight obstacles (three example environments with different difficulties are shown in Fig. \ref{fig::barn}). We randomly pick 250 environments for training and use the remaining 50 as the test set. In each of the environments, the robot aims to navigate from a fixed start to a fixed goal location in a safe and fast manner. 
For the reward function, while $R_f$ penalizes each time step before reaching the global goal, we simplified $R_p$ by replacing it with its projection along the $y$-axis (Fig. \ref{fig::barn}). The distance function in $d(o_{t+1})$ in $R_c$ is the minimal value among the 720 laser beams. $\pi_p$ produces a new set of planner parameters every two seconds. 

\begin{figure}
  \centering
  \includegraphics[width=\columnwidth]{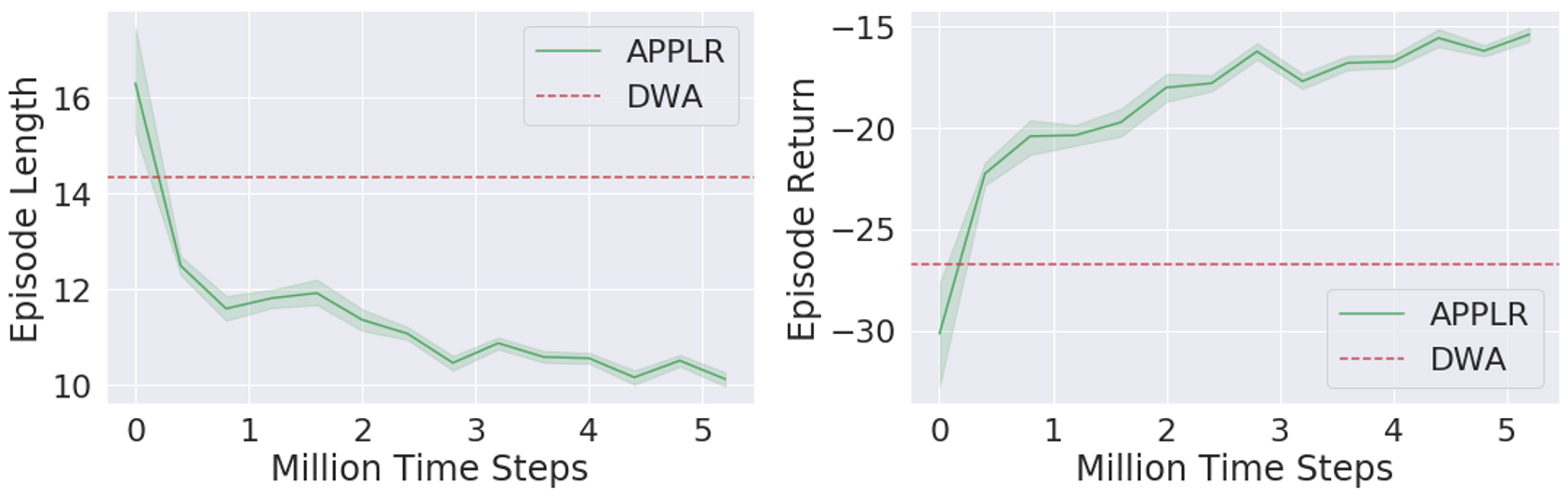}
  \caption{Learning Curves of Episode Length and Episode Return Averaged over 100 Episodes: The values are moving averaged over 40k steps. The red dashed lines mark the average performance of \textsc{dwa} planner with a static set of default parameters. The curves show significant improvement in the episode return and the time steps required to finish a navigation task.} 
  \label{fig::learning_curve}
\end{figure}

\begin{figure*}
  \centering
  \includegraphics[width=2\columnwidth]{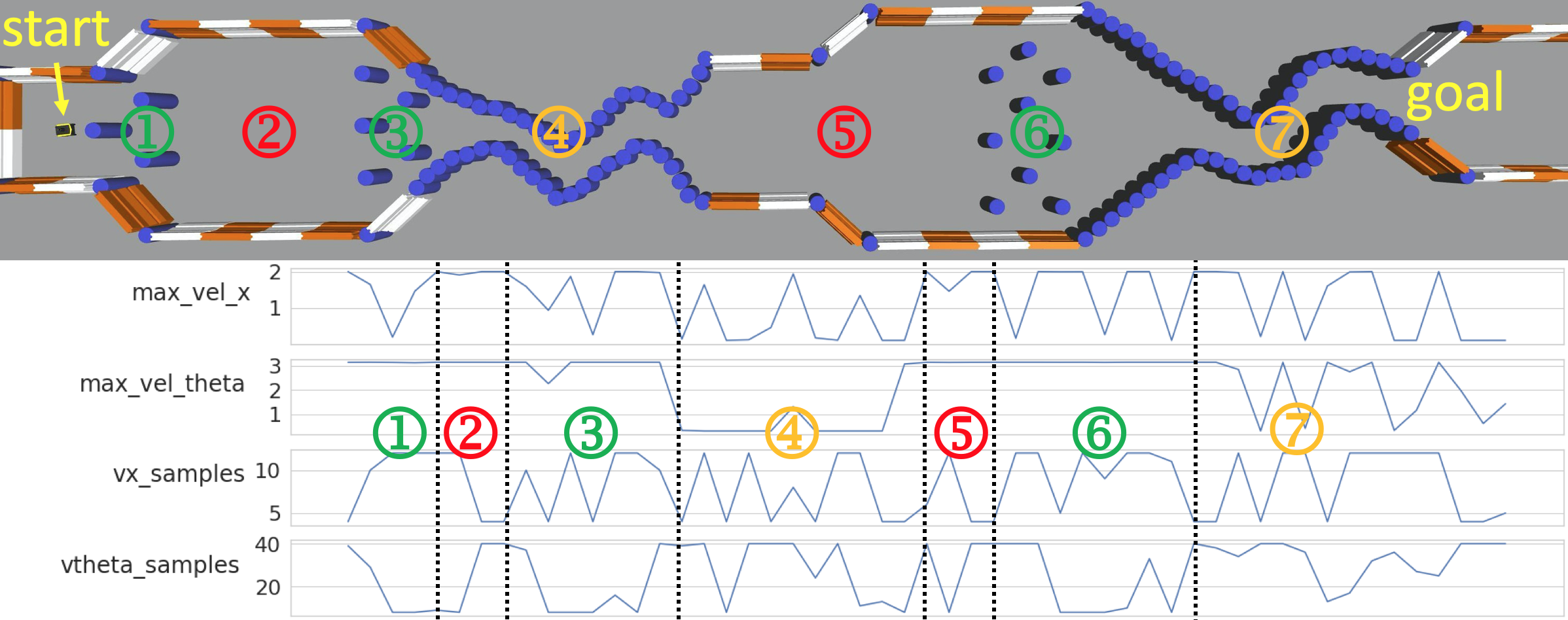}
  \caption{Example Parameter Profiles Selected by \textsc{applr}: Labels of qualitatively similar regions are in the same color. }
  \label{fig::gazebo_profile}
\end{figure*}

This simulated navigation task is implemented in a Singularity container, which enables easy parallelization on a computer cluster. 
TD3~\cite{fujimoto2018addressing} is implemented to learn the parameter policy $\pi_p$ in simulation. The policy network and the two target Q-networks are represented as multilayer perceptrons with three 512-unit hidden layers. The policy is learned under the distributed architecture Gorila~\cite{nair2015massively}. The acting processes are distributed over 500 CPUs with each CPU running one individual navigation task. On average, two actors work on a given navigation task. A single central learner periodically collects samples from the actors' local buffers and supplies the updated policy to each actor. Gaussian exploration noise with 0.5 standard deviation is applied to the actions at the beginning of the training. Afterward the standard deviation linearly decays at a rate of 0.125 per million steps and stays at 0.02 after 4 million steps. The entire training process takes about 6 hours and requires 5 million transition samples in total. Fig. \ref{fig::learning_curve} shows episode length and return averaged over 100 episodes and compares with \textsc{dwa} motion planner with default parameters. The episode return continually increases and episode length drops by around 40\% by the end of the training. As shown in Fig. \ref{fig::learning_curve}, \textsc{applr} surpasses \textsc{dwa} at an early stage of the training process in terms of both episode length and return. 
After training, we deploy the learned parameter policy $\pi_p$ in an example environment and plot four example parameter profiles produced by $\pi_p$ in Fig. \ref{fig::gazebo_profile}. Dashed lines separate different parts of the profile, which correspond to different regions in the example environment. 

\subsection{Simulated Experiments} 
\begin{figure}
     \centering
     \subfloat[Train Environments]{\includegraphics[width=0.46\columnwidth]{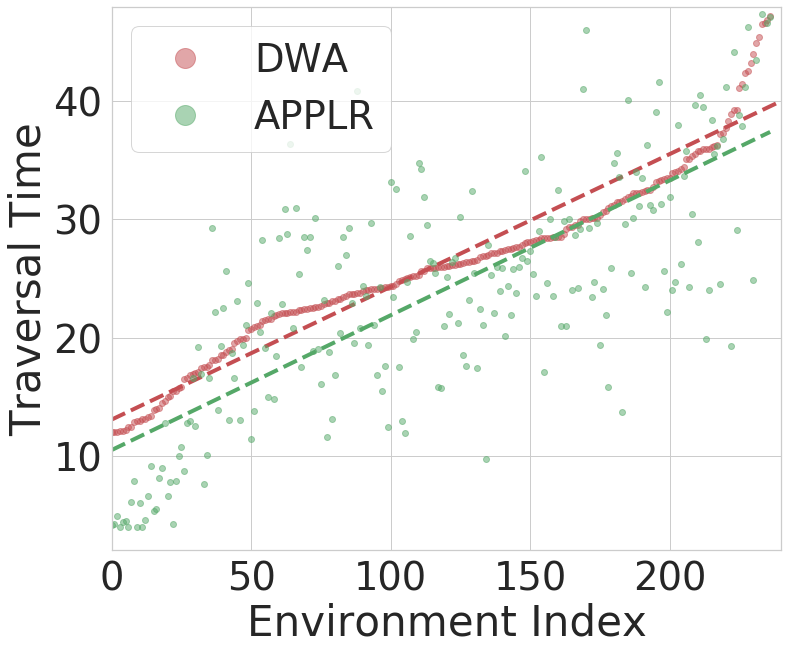}
     \label{fig::train_plot}}
     \hfill
     \subfloat[Test Environments]{\includegraphics[width=0.48\columnwidth]{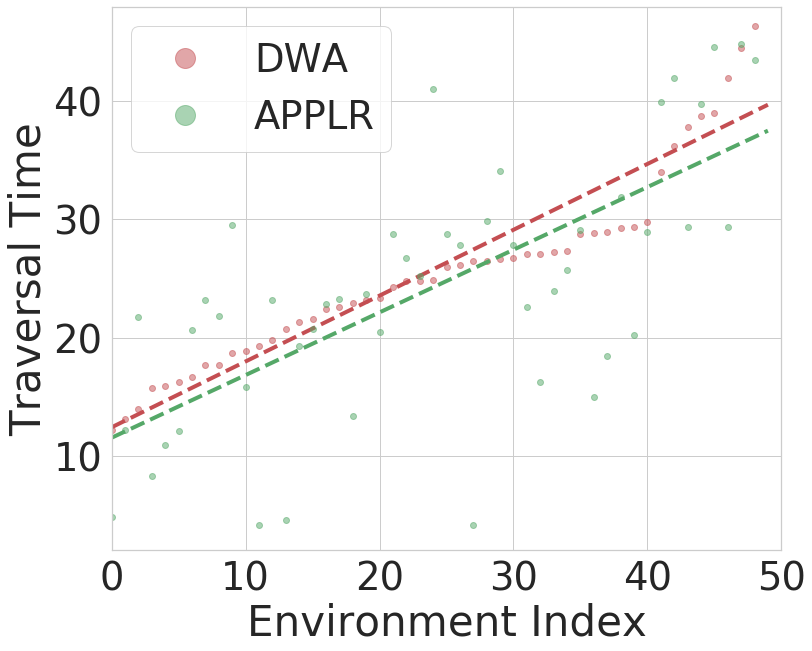}
     \label{fig::test_plot}}
    \caption{Traversal Time of \textsc{applr} and \textsc{dwa} in Training (a) and Test (b) Environments: The environment index is ordered by \textsc{dwa}'s traversal time, indicating difficulty level. Dashed lines represent linear fittings of traversal time vs. index. On average, \textsc{applr} achieves faster traversal than \textsc{dwa}.
    } 
    \label{fig:two_plots}
\end{figure}

After training, we deploy the learned parameter policy $\pi_p$ on both the training and test environments. We average over the traversal time of 20 trials for \textsc{dwa} and \textsc{applr} in each environment. The results over the 250 training environments are shown in Fig. \ref{fig::train_plot} and over the 50 test environments in Fig. \ref{fig::test_plot}. The majority of the green dots (\textsc{applr}) are beneath the red dots (\textsc{dwa}), indicating better performance by \textsc{applr}. We use linear regression to fit two lines for better visualization.
Tab. \ref{tab::average_epl} shows the average traversal time of \textsc{applr} and \textsc{dwa} and relative improvement. \textsc{applr} yields an improvement of 9.6\% in the training environments, and 6\% in the test environments. 

\begin{table}
\centering
\caption{Average Traversal Time of \textsc{applr} and \textsc{dwa}}
\begin{tabular}{ccccc}
\toprule
  & \textsc{applr} & \textsc{dwa} & Improvement &  \\ \midrule
Training & 11.96 & 13.24 & 1.28 (9.6\%)&   \\ 
Test & 12.25 & 13.03 & 0.78 (6\%) &  \\
\bottomrule
\end{tabular}
\label{tab::average_epl}
\vspace{-15pt}
\end{table}

To test the statistical significance of our simulation results, we run a t-test for each pair of \textsc{applr} and \textsc{dwa} performance in each environment. Tab. \ref{tab::t_test} compares the number of navigation environments where statistically significantly better navigation performance is achieved. 
In both training and test set, the results show that \textsc{applr} achieves statistically significantly better navigation performance in over 30\% of environments than \textsc{dwa} does, while \textsc{dwa} is only better in 9\% and 6\% of environments in the training and test set, respectively. 

\begin{table}
\centering
\caption{Number and Percentage of All Environments 
in which One Method is Better Compared to the Other
} 
\begin{tabular}{ccccc}
\toprule
  & \textsc{applr} better & \textsc{dwa} better &   \\ \midrule
Training & 88 ($35\%$) & 23 ($9\%$) &   \\ 
Test & 15 (30\%) & 3 (6\%) &   \\
\bottomrule
\end{tabular}
\label{tab::t_test}
\vspace{-10pt}
\end{table}

Furthermore, we analyze the relationship between performance improvement and difficulty level of a particular environment. We classify the first one third of environments (100) where \textsc{dwa} achieves the fastest traversal times as Easy, the one third of environments (100) with slowest traversal times as Difficult, and the remaining one third of environments (100) as Medium (Tab. \ref{tab::difficulties}). While the advantage of \textsc{applr} over \textsc{dwa} is evident for the Easy environments (57\% vs. 8\%), it diminishes with increased environment difficulty (Medium: 24\% vs. 3\% and Difficult: 28\% vs. 14\%). We conjecture that this relationship may be due to a potential performance upper bound of the \textsc{dwa} planner due to its underlying structure. That is, while selecting the right planner parameters at each time step in easy environments can significantly improve its performance, in difficult environments, \textsc{dwa}'s performance has saturated such that selecting the right parameters can only lead to marginal improvement. In those environments, it is likely that a completely different planner is required to achieve better performance, e.g. an end-to-end planner~\cite{xiao2020toward}.

\begin{table}
\centering
\caption{Percentage of Environments (in which one method is better compared to the other) under Different Difficulty Levels 
}
\begin{tabular}{lcccc}
\toprule
  & \textsc{applr} better & \textsc{dwa} better &   \\ \midrule
Easy Train& 62\% &  7\%  &   \\
Easy Test & 41\% & 12\% &   \\ 
\textbf{Easy All} & \textbf{57\%} &  \textbf{8\%}  &   \\\midrule
Medium  Train & 29\% & 14\% &   \\
Medium Test & 18\% & 6\% &   \\ 
\textbf{Medium All} & \textbf{24\%} & \textbf{3}\% &   \\\midrule
Difficult Train & 26\% & 5\% &   \\
Difficult Test & 31\% & 0\% &   \\
\textbf{Difficult All} & \textbf{28\%} & \textbf{14\%} &   \\
\bottomrule
\end{tabular}
\label{tab::difficulties}
\vspace{-20pt}
\end{table}

\subsection{Physical Experiments}
\begin{figure}
  \centering
  \includegraphics[width=\columnwidth]{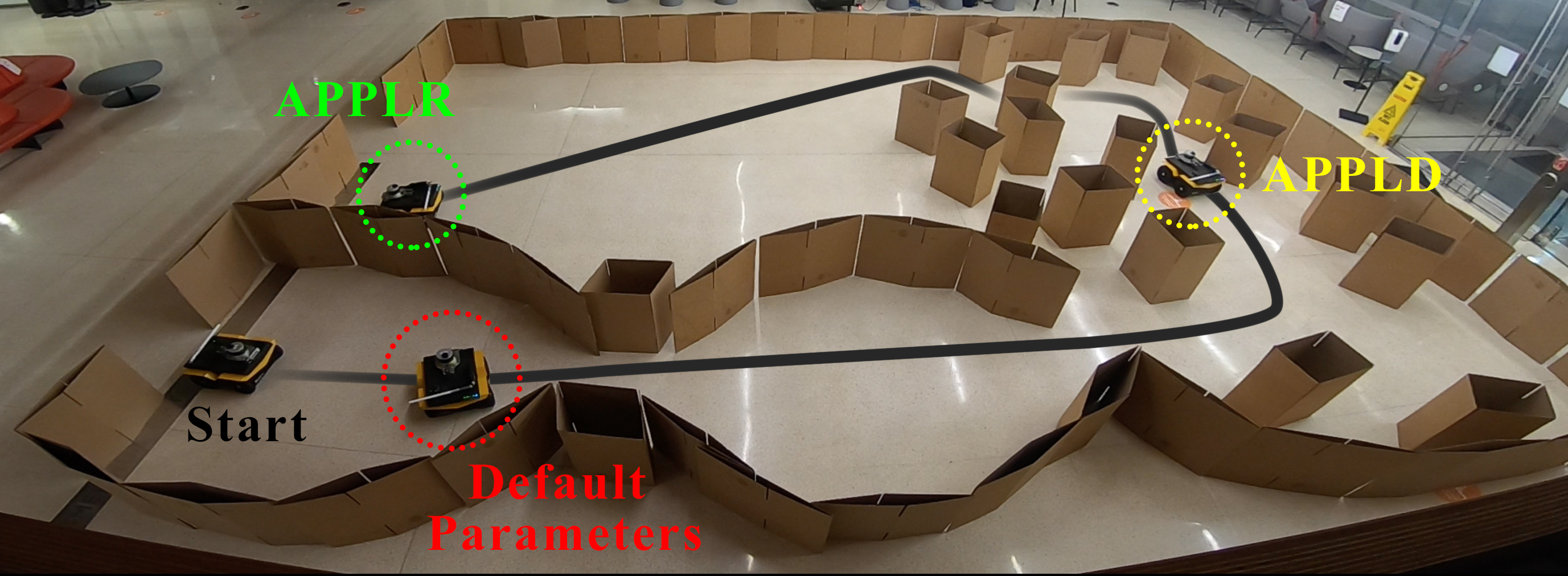}
  \caption{Physical Experiments: While the \textsc{dwa} planner with a static set of default parameters (red) fails to find feasible motions and executes recovery behaviors in many places, \textsc{appld} (yellow) and \textsc{applr} (red) can both successfully and smoothly navigate through the entire obstacle course. Using RL, \textsc{applr} can achieve faster traversal than \textsc{appld} learned from (most likely suboptimal) human demonstration. }
  \label{fig::physical}
\end{figure}

To validate the sim-to-real transfer of \textsc{applr}, we also test the learned parameter policy $\pi_p$ on a physical Jackal robot. The physical robot has a Velodyne LiDAR, but we transform the 3D point cloud to the same 2D laser scan as in the simulation (720-dimensional and 270$^\circ$ field of view). The learned policy is deployed in a real-world obstacle course set up by cardboard boxes (Fig. \ref{fig::physical}). This physical environment is different from any of the navigation environments in the BARN dataset. Therefore, both generalizability and sim-to-real transfer of \textsc{applr} can be tested with this unseen real-world environment. 

Given this target environment, we further collect a teleoperated demonstration provided by one of the authors and learn a parameter tuning policy based on the notion of navigational context (\textsc{appld}~\cite{xiao2020appld}). The author aims at driving the robot to traverse the entire obstacle course in a safe and fast manner. \textsc{appld} identifies three contexts using the human demonstration and learns three sets of navigation parameters. 

We compare the performance of \textsc{applr} with that of \textsc{appld} and the \textsc{dwa} planner using a set of hand-tuned default parameters. For each trial, the robot navigates from the fixed start point to a fixed goal point. Each trial is repeated five times and we report the mean and standard deviation in Tab. \ref{tab::physical_results}. 
We observe one failure trial (the robot fails to find feasible motions and keeps rotating in place at the beginning of the narrow part) with \textsc{dwa}. Therefore, the \textsc{dwa} results only contain the four successful trials.


\begin{table}
\centering
\vspace{-7pt}
\caption{Traversal Time in Physical Experiments \\ \centering \small{(* denotes one additional failure trial)}}
\begin{tabular}{ccc}
\toprule
\textsc{dwa} & \textsc{appld} & \textsc{applr} \\ 
\midrule
72.8$\pm$10.1s* & 43.2$\pm$4.1s  & \textbf{34.4$\pm$4.8s}\\
\bottomrule
\end{tabular}
\label{tab::physical_results}
\vspace{-12pt}
\end{table}

In all \textsc{dwa} trials, the robot gets stuck in many places, especially where the surrounding obstacles are very tight. It has to engage in many recovery behaviors, i.e. rotating in place or driving backwards, to ``get unstuck''. Furthermore, in relatively open space, the robot drives unnecessarily slowly. All these behaviors contribute to the large traversal time and high variance (plus an additional failure trial). 
Unlike many simulation environments in BARN~\cite{perille2020benchmarking}, where obstacles are generated by cellular automata and therefore very cluttered, the relatively open space in the physical environment (Fig. \ref{fig::physical}) allows faster speed and gives \textsc{applr} a greater advantage.  
Surprisingly, \textsc{applr} even achieves better navigation performance than \textsc{appld}, which has access to a human demonstration in the same environment. One of the reasons we observe this result in the physical experiments is that the human demonstrator is relatively conservative in some places; the parameters learned by \textsc{appld} are upper-bounded by this suboptimal human performance. On the other hand, the RL parameter policy aims at reducing the traversal time and finds better parameter sets to achieve that goal. Another reason is that while \textsc{appld} only utilizes three sets of learned parameters for the three navigational contexts, \textsc{applr} is given the flexibility to change parameters at each time step, and RL is able to utilize the sequential aspect of the parameter selection problem, e.g. slowing down in order to speed up in the future. However, we observe that in confined spaces \textsc{applr} produces less smooth motion compared to \textsc{appld}. One possible explanation is that the teleoperated human demonstration in \textsc{appld} aims at both fast and smooth navigation, while \textsc{applr} only aims at speed. This issue may be addressable in future work through a different reward function. 
\section{CONCLUSIONS}
\label{sec::conclusions}

In this paper, we introduce \textsc{applr}, \emph{Adaptive Planner Parameter planning from Reinforcement}, which, in contrast to \emph{parameter tuning}, learns a \emph{parameter policy}.
The parameter policy is trained using RL to select planner parameters at each time step to allow the robot to take suboptimal actions in a current state in order to achieve better future performance. 
Furthermore, instead of learning an end-to-end navigation planner, we treat a classical motion planner as part of the environment and the RL agent only interacts with it through the planner parameters. Learning in this parameter space instead of a velocity control space not only allows \textsc{applr} to inherit all the benefits of classical navigation systems, such as safety and explainability, but also eliminates wasteful random exploration with the help of the underlying planner, allows it to generalize well to unseen environments, and reduces the chances of failing to overcome the sim-to-real transfer gap. The unsupervised \textsc{applr} paradigm does not require any expert tuning or human demonstration. \textsc{applr} is trained on a suite of simulated navigation environments and is then tested in unseen environments. We also conduct physical experiments to test \textsc{applr}'s sim-to-real transfer. 
As mentioned in Sec. \ref{sec::experiments}, one interesting direction for future work is to design reward functions that encourage motion smoothness. 
Currently, \textsc{applr} is only useful if there’s no switching cost in the planner for changing parameter sets, but, if such a cost exists, future work should take it into account. 
Another interesting direction is to use curriculum learning to start from easy environments and then transition to difficult ones. Furthermore, the \textsc{applr} pipeline has the potential to be applied to other navigation systems, including visual, semantic, or aerial navigation. 


\bibliographystyle{IEEEtran}
\bibliography{IEEEabrv,references}

\begin{thebibliography}{10}
\providecommand{\url}[1]{#1}
\csname url@samestyle\endcsname
\providecommand{\newblock}{\relax}
\providecommand{\bibinfo}[2]{#2}
\providecommand{\BIBentrySTDinterwordspacing}{\spaceskip=0pt\relax}
\providecommand{\BIBentryALTinterwordstretchfactor}{4}
\providecommand{\BIBentryALTinterwordspacing}{\spaceskip=\fontdimen2\font plus
\BIBentryALTinterwordstretchfactor\fontdimen3\font minus
  \fontdimen4\font\relax}
\providecommand{\BIBforeignlanguage}[2]{{%
\expandafter\ifx\csname l@#1\endcsname\relax
\typeout{** WARNING: IEEEtran.bst: No hyphenation pattern has been}%
\typeout{** loaded for the language `#1'. Using the pattern for}%
\typeout{** the default language instead.}%
\else
\language=\csname l@#1\endcsname
\fi
#2}}
\providecommand{\BIBdecl}{\relax}
\BIBdecl

\bibitem{zheng2017ros}
K.~Zheng, ``Ros navigation tuning guide,'' \emph{arXiv preprint
  arXiv:1706.09068}, 2017.

\bibitem{xiao2017uav}
X.~Xiao, J.~Dufek, T.~Woodbury, and R.~Murphy, ``Uav assisted usv visual
  navigation for marine mass casualty incident response,'' in \emph{2017
  IEEE/RSJ International Conference on Intelligent Robots and Systems
  (IROS)}.\hskip 1em plus 0.5em minus 0.4em\relax IEEE, 2017, pp. 6105--6110.

\bibitem{xiao2020appld}
X.~Xiao, B.~Liu, G.~Warnell, J.~Fink, and P.~Stone, ``Appld: Adaptive planner
  parameter learning from demonstration,'' \emph{IEEE Robotics and Automation
  Letters}, vol.~5, no.~3, pp. 4541--4547, 2020.

\bibitem{teso2019predictive}
D.~Teso-Fz-Beto{\~n}o, E.~Zulueta, U.~Fernandez-Gamiz, A.~Saenz-Aguirre, and
  R.~Martinez, ``Predictive dynamic window approach development with artificial
  neural fuzzy inference improvement,'' \emph{Electronics}, vol.~8, no.~9, p.
  935, 2019.

\bibitem{bhardwaj2019differentiable}
M.~Bhardwaj, B.~Boots, and M.~Mukadam, ``Differentiable gaussian process motion
  planning,'' \emph{arXiv preprint arXiv:1907.09591}, 2019.

\bibitem{pfeiffer2017perception}
M.~Pfeiffer, M.~Schaeuble, J.~Nieto, R.~Siegwart, and C.~Cadena, ``From
  perception to decision: A data-driven approach to end-to-end motion planning
  for autonomous ground robots,'' in \emph{{IEEE International Conference on
  Robotics and Automation}}.\hskip 1em plus 0.5em minus 0.4em\relax IEEE, 2017.

\bibitem{gao2017intention}
W.~Gao, D.~Hsu, W.~S. Lee, S.~Shen, and K.~Subramanian, ``Intention-net:
  Integrating planning and deep learning for goal-directed autonomous
  navigation,'' \emph{arXiv preprint arXiv:1710.05627}, 2017.

\bibitem{chiang2019learning}
H.-T.~L. Chiang, A.~Faust, M.~Fiser, and A.~Francis, ``Learning navigation
  behaviors end-to-end with autorl,'' \emph{IEEE Robotics and Automation
  Letters}, vol.~4, no.~2, pp. 2007--2014, 2019.

\bibitem{xiao2020toward}
X.~Xiao, B.~Liu, G.~Warnell, and P.~Stone, ``Toward agile maneuvers in highly
  constrained spaces: Learning from hallucination,'' \emph{arXiv preprint
  arXiv:2007.14479}, 2020.

\bibitem{liu2020lifelong}
B.~Liu, X.~Xiao, and P.~Stone, ``Lifelong navigation,'' \emph{arXiv preprint
  arXiv:2007.14486}, 2020.

\bibitem{xiao2020agile}
X.~Xiao, B.~Liu, and P.~Stone, ``Agile robot navigation through hallucinated
  learning and sober deployment,'' \emph{arXiv preprint arXiv:2010.08098},
  2020.

\bibitem{siva2019robot}
S.~Siva, M.~Wigness, J.~Rogers, and H.~Zhang, ``Robot adaptation to
  unstructured terrains by joint representation and apprenticeship learning,''
  in \emph{Robotics: Science and Systems (RSS)}, 2019.

\bibitem{kahn2020badgr}
G.~Kahn, P.~Abbeel, and S.~Levine, ``Badgr: An autonomous self-supervised
  learning-based navigation system,'' \emph{arXiv preprint arXiv:2002.05700},
  2020.

\bibitem{everett2018motion}
M.~Everett, Y.~F. Chen, and J.~P. How, ``Motion planning among dynamic,
  decision-making agents with deep reinforcement learning,'' in \emph{2018
  IEEE/RSJ International Conference on Intelligent Robots and Systems
  (IROS)}.\hskip 1em plus 0.5em minus 0.4em\relax IEEE, 2018, pp. 3052--3059.

\bibitem{pokle2019deep}
A.~Pokle, R.~Mart{\'\i}n-Mart{\'\i}n, P.~Goebel, V.~Chow, H.~M. Ewald, J.~Yang,
  Z.~Wang, A.~Sadeghian, D.~Sadigh, S.~Savarese \emph{et~al.}, ``Deep local
  trajectory replanning and control for robot navigation,'' in \emph{2019
  International Conference on Robotics and Automation (ICRA)}.\hskip 1em plus
  0.5em minus 0.4em\relax IEEE, 2019, pp. 5815--5822.

\bibitem{fujimoto2018addressing}
S.~Fujimoto, H.~van Hoof, and D.~Meger, ``Addressing function approximation
  error in actor-critic methods,'' 2018.

\bibitem{silver2014deterministic}
D.~Silver, G.~Lever, N.~Heess, T.~Degris, D.~Wierstra, and M.~Riedmiller,
  ``Deterministic policy gradient algorithms,'' 2014.

\bibitem{hasselt2010double}
H.~V. Hasselt, ``Double q-learning,'' in \emph{Advances in neural information
  processing systems}, 2010, pp. 2613--2621.

\bibitem{nair2015massively}
A.~Nair, P.~Srinivasan, S.~Blackwell, C.~Alcicek, R.~Fearon, A.~D. Maria,
  V.~Panneershelvam, M.~Suleyman, C.~Beattie, S.~Petersen, S.~Legg, V.~Mnih,
  K.~Kavukcuoglu, and D.~Silver, ``Massively parallel methods for deep
  reinforcement learning,'' 2015.

\bibitem{fox1997dynamic}
D.~Fox, W.~Burgard, and S.~Thrun, ``The dynamic window approach to collision
  avoidance,'' \emph{IEEE Robotics \& Automation Magazine}, vol.~4, no.~1, pp.
  23--33, 1997.

\bibitem{perille2020benchmarking}
D.~Perille, A.~Truong, X.~Xiao, and P.~Stone, ``Benchmarking metric ground
  navigation,'' \emph{arXiv preprint arXiv:2008.13315}, 2020.

\end{thebibliography}

\end{document}